%% file: main_hu.tex
\documentclass[11pt]{article}
\usepackage{styles/seed}
\AtBeginDocument{%
  \let\origfigure\figure\let\endorigfigure\endfigure
  \renewenvironment{figure}[1][]{\origfigure[htbp]}{\endorigfigure}%
  \let\origtable\table\let\endorigtable\endtable
  \renewenvironment{table}[1][]{\origtable[htbp]}{\endorigtable}%
  \renewenvironment{figure*}[1][]{\origfigure[htbp]}{\endorigfigure}%
  \renewenvironment{table*}[1][]{\origtable[htbp]}{\endorigtable}}
\usepackage{latexsym}\usepackage{amsmath}\usepackage{amssymb}
\usepackage{booktabs}\usepackage[T1]{fontenc}\usepackage[utf8]{inputenc}
\usepackage{microtype}\usepackage{graphicx}\usepackage[table]{xcolor}
\usepackage{enumitem}
\usepackage{todonotes}\usepackage{capt-of}\usepackage{adjustbox}\usepackage{float}
\newcommand{\cnum}[1]{\textcircled{\scriptsize #1}}
\definecolor{accent}{HTML}{B8562B}   % shortfall against full precision (appendix grids)
\definecolor{good}{HTML}{3C7D5A}     % score retained against full precision
\definecolor{cost}{HTML}{2E6E8E}     % memory a row holds

\title{PACE: Periodic and Channel-Aligned Encoding\\of Four-Bit Recurrent States}
\author{Patrick Haller \quad Filipe Laitenberger \quad Alan Akbik}
\affiliations{Humboldt-Universit\"at zu Berlin}
\preprintlogo{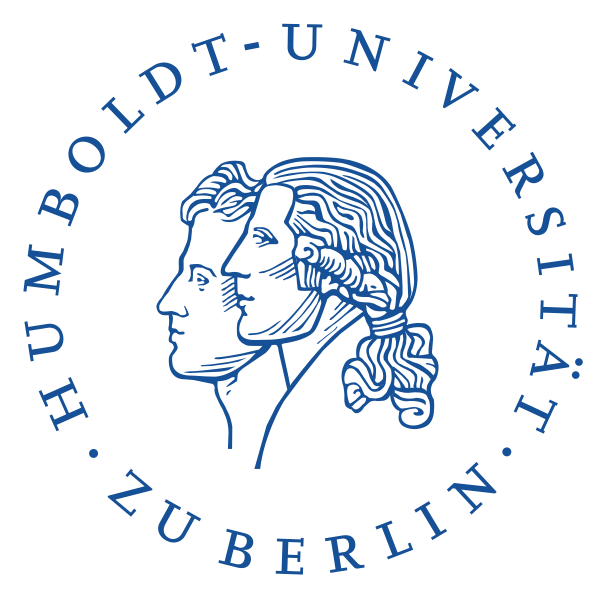}
\abstractfilledfalse
\preprintdate{\today}
\correspondence{\texttt{patrick.haller.1@hu-berlin.de}}
\newcommand{\colfigwidth}{0.6\linewidth}

\begin{document}
\maketitle

\begin{abstract}
Linear-attention language models keep their context in a recurrent state of fixed size, where an inference system stores one copy of this state for every sequence it decodes at the same time. Storing the state at lower precision therefore lets more sequences share an accelerator. Existing state quantizers round the state after every generated token and report that four bits lose accuracy unless the bit width is raised or the model is fine-tuned.\par We introduce PACE, a post-hoc four-bit state quantizer. PACE rounds the state only once every few generated tokens, as chunked kernels already do once per chunk during prefill, and keeps the tokens generated in between in a small buffer. Rounding less often reduces the error that accumulates in the state. Additionally, we find that the large entries of the gated-delta state gather in columns, because some channels of the update written at each token remain large across many tokens. PACE therefore shares each quantization scale within one column. It also keeps the heads that retain rounding error longest at eight bits. On four hybrid models from 4B to 35B parameters, PACE with mixed four- and eight-bit heads is competitive with eight-bit DAMP on the three larger models at about two thirds of its memory, PACE with four-bit heads throughout scores higher than Q-Mamba at equal memory, and rounding every few tokens preserves the decoding speedup of the buffer.
\end{abstract}

\begin{figure}[t]
\centering
% Preprint: the figure at its drawn width (3.3in), so it fits on page 1.
\includegraphics[width=0.5\linewidth]{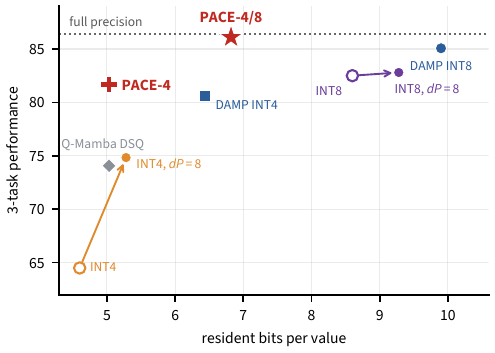}
\caption{Mean performance of MMLU-Pro, IFEval and RULER-32k against resident bits per value on Qwen3.5-9B.
Arrows join per-token and $dP{=}8$ rounding variants.}
\label{fig:frontier}
\end{figure}

\section{Introduction}
\label{sec:intro}
Linear-attention language models replace the key--value cache of a Transformer with a recurrent
state of fixed size and have become competitive with Transformers \citep{yang2024gla,yang2025gateddeltanet}.
Every recurrent layer holds one state matrix per head, and a server stores one copy for every
sequence it decodes at the same time. At 64 concurrent sequences, the fp32 states take between 10
and 41\,\% of the memory that weights and states occupy together. Storing the state at lower
precision lets more sequences fit on the same accelerator.
Existing state quantizers \citep{chiang2025quamba2,chen2025qmamba,zhang2026damp,nvidia2026nemotron}
differ in how they scale and round the state, and all of them round it after every generated token.
With per-token rounding, the methods that evaluate a four-bit state report a loss in accuracy (Figure~\ref{fig:frontier}). The published remedies either
raise the bit width to eight bits or fine-tune part of the model weights. \citet{nvidia2026ultra} round the state less often, but only with eight or more bits per value.

\begin{figure*}[b!]
\includegraphics[width=\textwidth]{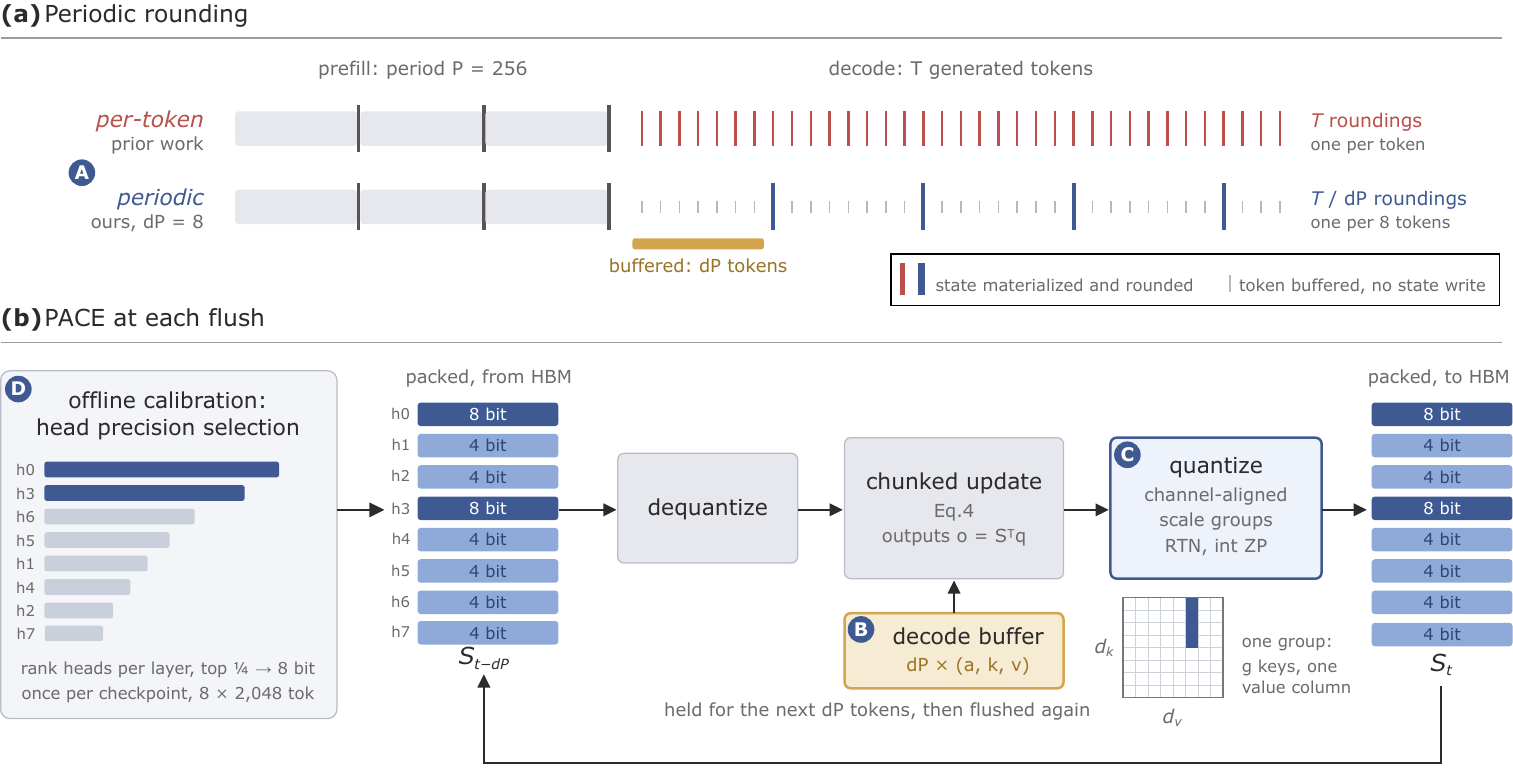}
\caption{Overview of PACE. (a) Prior state quantizers round the recurrent state after every generated
token; PACE rounds it once every $dP$ tokens (\cnum{A}), as prefill already does at each kernel chunk
boundary. (b) At each flush, the packed state is dequantized, advanced over the tokens held in the
decode buffer (\cnum{B}), and requantized in channel-aligned scale groups (\cnum{C}). PACE-4/8 also holds a quarter of the heads
in each layer at eight bits, selected once per checkpoint (\cnum{D}).}
\label{fig:method_overview}
\end{figure*}

During prefill, chunked kernels process $64$ to $1{,}024$ tokens at a time
and write the state only at the end of each chunk, so every published quantizer already rounds the
prefill state once per chunk. During decoding the kernel is called once per generated token, so the
state is written, and rounded, after every token (Figure~\ref{fig:method_overview}a).

In this work, we introduce PACE (Periodic And Channel-aligned state Encoding), a post-hoc four-bit
state quantizer that requires no training (Figure~\ref{fig:method_overview}). During decoding, PACE
rounds the state once every $dP$ generated tokens (the decode period) instead of after every token (\cnum{A}). The tokens
generated in between are held in a decode buffer, a small store of recent keys and values from which
the state is advanced in one step. Recent serving work uses such buffers to speed up decoding
\citep{zou2026kvbuffer,daolab2026replayssm,lin2026deltalog}. The buffer occupies memory, and
every memory total we report counts it together with the state (\cnum{B}).

The gated-delta state has one row per key dimension and one column per channel of the delta value,
the vector the recurrence writes into the state at each token. Channels that stay large across many
tokens produce columns of large entries, and when the channels are permuted at every token, the large
entries no longer form columns (Section~\ref{sec:structure}). PACE therefore forms each scale group
from the entries of one column, that is, along the key axis (\cnum{C}). PACE has two variants:
PACE-4 keeps all heads at four bits, and PACE-4/8 keeps the quarter of heads in each layer that retain
rounding error longest at eight bits (\cnum{D}).

Our contributions are:
\begin{itemize}[leftmargin=*,itemsep=1pt,parsep=0pt,topsep=2pt]
\item A bound on the rounding error a head retains as a function of the rounding period and the
head's decay (Section~\ref{sec:period}), together with a memory accounting that includes the decode
buffer (Section~\ref{sec:method:cost}). On four backbones, rounding every eight generated tokens
reduces the accuracy loss at every bit width below eight (Section~\ref{sec:res:period}).
\item The observation that the large entries of the gated-delta state form columns, explained by
channels of the delta value that stay large across tokens and confirmed by permuting those channels (Section~\ref{sec:structure}).
\item PACE, a four-bit state quantizer with channel-aligned scale groups and head-level precision
allocation (Section~\ref{sec:method}), compared with DAMP and Q-Mamba on four hybrid models from 4B to
35B parameters (Section~\ref{sec:headline}).
\item Triton kernels for periodic decoding with a packed four-bit state, with which we show that
rounding every few tokens keeps the decoding speedup of the decode buffer (Section~\ref{sec:res:thru}).
\end{itemize}

\newpage
\section{Background and Related Work}
\label{sec:bg}
\label{sec:rel}

\subsection{Gated Delta Rule}
\label{sec:bg:state}
Linear attention \citep{katharopoulos2020linear,schlag2021fwp} removes the softmax, so that attention
becomes a recurrence over a state $S_t \in \mathbb{R}^{d_k \times d_v}$ per head,
\begin{equation}
S_t = S_{t-1} + k_t v_t^\top , \qquad o_t = S_t^\top q_t .
\label{eq:la}
\end{equation}
The state replaces the key--value cache. It holds the whole past, does not grow with context, and is read
and written at every step.

Variants differ in how $S_{t-1}$ is modified before the new token is written
\citep{sun2023retnet,peng2023rwkv,yang2024gla,beck2024xlstm}. We work with the gated delta rule
\citep{yang2025gateddeltanet}, which applies a data-dependent decay $a_t \in (0,1)$ and removes the
component of the state along the current key before writing,
\begin{equation}
S_t = a_t\left(I - \beta_t k_t k_t^\top\right) S_{t-1} + \beta_t k_t v_t^\top ,
\label{eq:gdr}
\end{equation}
with write strength $\beta_t$. The read-out is unchanged. Expanding Eq.~\ref{eq:gdr}, every write is an
outer product,
\begin{equation}
S_t = a_t S_{t-1} + k_t u_t^\top, \;\; u_t = \beta_t (v_t - a_t S_{t-1}^\top k_t),
\label{eq:delta}
\end{equation}
where we call $u_t$ the \emph{delta value}, the part of $v_t$ that the decayed state does not already
return for $k_t$, scaled by $\beta_t$.

\subsection{Chunk-wise Kernels}
\label{sec:bg:chunk}
Evaluating Eq.~\ref{eq:gdr} token by token is sequential and memory-bound. Chunk-wise kernels
\citep{yang2024gla,yang2025gateddeltanet} split the sequence into chunks of length $C$ and process each
chunk from the state at its start. Writing $\alpha_i = \prod_{j \le i} a_j$ for the cumulative decay
within a chunk, the state after $C$ tokens is
\begin{equation}
S_C = \alpha_C S_0 + \sum_{i=1}^{C} \frac{\alpha_C}{\alpha_i}\, k_i u_i^\top ,
\label{eq:chunk}
\end{equation}
and the outputs follow from $S_0$ and the chunk's own tokens by matrix multiplication. The delta values
$u_i$ of Eq.~\ref{eq:delta} are obtained from a triangular solve over the chunk, so Eq.~\ref{eq:chunk} is evaluated
with dense arithmetic rather than a loop. Only $S_0$ and $S_C$ are formed as tensors. The states at
positions inside the chunk are never written.

\begin{figure*}[h]
\centering
\includegraphics[width=\textwidth]{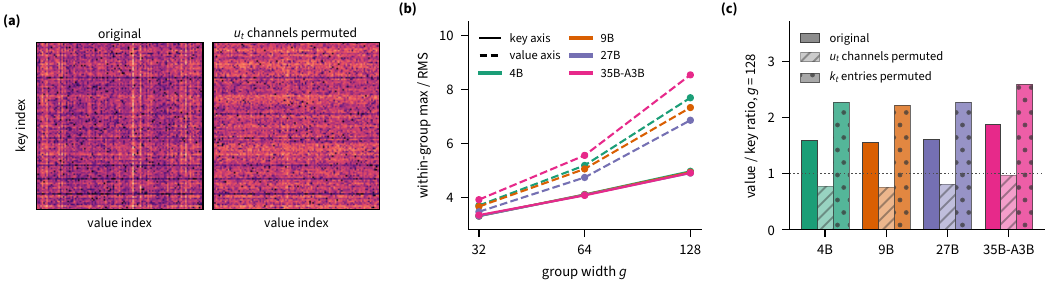}
\caption{Large entries of the recurrent state form columns indexed by the value coordinate.
(a) $|S|$ of one head of Qwen3.5-9B on a shared logarithmic scale, in the original model and with the channels of
the delta value $u_t$ permuted at every token; the head is the median head of the median layer by the
value-to-key ratio. (b) Within-group max/RMS against group width, for groups along the key axis
(solid) and the value axis (dashed). (c) Ratio of the value-axis to
the key-axis statistic at $g{=}128$, in the original model and with $u_t$ or $k_t$ permuted at every token.}
\label{fig:axis}
\end{figure*}

\subsection{Decode Buffers}
\label{sec:bg:buffer}
Decoding calls the kernel once per token, so the state is written every step and the chunk structure is
lost. Recent serving work recovers it with a decode buffer. Recent keys and values are held, and the state is
advanced once per flush, the step at which the buffered tokens are applied, instead of once per token.
This removes the per-token write of the state from memory traffic.
KVBuffer \citep{zou2026kvbuffer} derives an optimal buffer of about $23$ tokens and reports its best
latency at $32$, ReplaySSM \citep{daolab2026replayssm} uses $16$ for Qwen3.5, and DeltaLog
\citep{lin2026deltalog} merges every $8$ tokens for Gated DeltaNet. None of them quantizes the state.

\subsection{State Quantization}
\label{sec:bg:quant}
\paragraph{Quantizing the recurrent state.}
Quamba2 \citep{chiang2025quamba2} stores cached states at eight bits with static scales. Q-Mamba
\citep{chen2025qmamba} reaches four bits by scaling the state with one vector per axis and fine-tuning
about 2\% of the weights. MambaQuant \citep{xu2025mambaquant} rotates weights and activations before
rounding and leaves the state unquantized. DAMP \citep{zhang2026damp} ranks key channels by quantization error times decay persistence, keeps the riskiest in fp16 and reports $9.9$ bits per
value; it rounds once per chunk in prefill and after every token in decoding. The Nemotron report
\citep{nvidia2026nemotron} keeps the state in fp16 with stochastic rounding, and \citet{nvidia2026ultra}
store the state every eight decoding steps at eight bits and above, in emulation. \citet{kozyrev2026gdn}
quantize the weights and activations of the recurrent layers and leave the state unquantized.
None of them varies the rounding period below eight bits or counts the buffer it requires.

\paragraph{Error propagation.}
Rounding a recurrence is the fixed-point filter problem \citep{jackson1970roundoff,oppenheim1972finite}, in which rounding
noise of variance $\sigma^2$, injected at every step of a recurrence with decay $a$, accumulates to
$\sigma^2/(1-a^2)$. \citet{zhang2026damp} rederive the
accumulated error for the gated delta rule and reduce the error per rounding; we reduce the number of
roundings (Eq.~\ref{eq:retained}). \citet{kozyrev2026gdn} find that whole-state error stays flat from token $256$ on. \citet{zhao2025smallssm} clip the state during quantization-aware training and
\citet{siegel2026qs4d} quantize an S4D state through its convolutional form.

\paragraph{KV-cache quantization.}
KIVI \citep{liu2024kivi} and KVQuant \citep{hooper2024kvquant} reach two to four bits on a cache whose
entries are written once and only read, and rotation methods \citep{chee2023quip,ashkboos2024quarot,
liu2025spinquant} redistribute magnitude before rounding. A recurrent state is instead re-rounded at
every write.

% body_draft.tex -- replacement for sections 4-8 of skeleton.tex (2026-10-05).
% LaTeX fragments only: \input this after section 3 of skeleton.tex, before \bibliography.
% Every number below is read from tabs/*.tex, runs/20260915-battery/{DECODE_SPLIT,SCALE_GRETE}.md,
% SURPRISES.md, or a result.json named in a comment. Holes are LaTeX comments beginning "HOLE:".
%
% Main-body figures (two):  fig:where   figs/fig2_where.pdf   (schematic, no measurements)
%                           fig:memory  figs/fig3_memory.pdf  (measured buffer cost; not schedule-dependent)
% Main-body tables (four):  tab:opening:4b (Section 1), tab:headline:4b, tab:headline, tab:alloc (below)
% Do NOT use fig1b_collapse, fig1_period, fig1_tradeoff, fig4_matched: all four encode the one-period
% schedule (a single period applied to prefill and decode alike) and must be regenerated from the
% split-schedule cells before they can appear anywhere.
% tab_exposure.tex and tab_period.tex likewise encode the one-period schedule; both are cited below only
% from the appendix and are flagged there.

\section{Rounding Period and Error Propagation}
\label{sec:period}

\subsection{The Rounding Period}
\label{sec:method:period}
A quantizer can round the state only where the kernel writes it, which gives three regimes under Eq.~\ref{eq:chunk}
(Figure~\ref{fig:method_overview}a). In prefill the chunked kernel forms $S_0$ as the initial state and $S_C$, so the state is rounded once per chunk without an additional kernel call. In unbuffered decoding each call carries one token,
$S_t$ is formed at every step, and the period is one. Under a decode buffer the state is advanced once per flush, and the period is the number of tokens the buffer holds.

We call one choice of bit width, scale groups, buffer precision and rounding periods for the
quantized state a \emph{configuration}. Each configuration carries two periods. We write $P$ for the prefill period, which equals the kernel chunk
length $C$ and which we fix at $256$ tokens in all benchmark evaluations, and $dP$ for the decode period, which is the variable under study.

\subsection{Error Propagation}
\label{sec:period:error}
Each rounding adds an error to the state, and the recurrence carries it forward. Between two
roundings the error is multiplied at every token by $a_t(I - \beta_t k_t k_t^\top)$
(Eq.~\ref{eq:gdr}), whose norm is at most $a_t$ for a unit-norm key. With $a_h$ the mean decay of head
$h$, an error of energy $e_h$ injected at one rounding retains at most $a_h^{2\,dP}$ of its energy at
the next, and independent errors from successive roundings accumulate to at most
\begin{equation}
E_h = \frac{e_h}{1 - a_h^{2\,dP}} .
\label{eq:retained}
\end{equation}
A longer period lowers $E_h$, because more decay separates two roundings. In the gated delta rule the decay is one scalar per head, so
$E_h$ is a per-head quantity.

\noindent\textbf{Finding 1.} The rounding error a head retains falls with the rounding period and
rises with the head's decay

\section{Channel Structure of the Recurrent State}
\label{sec:structure}
A group-wise quantizer divides the state of each head into scale groups of $g$ consecutive entries
and rounds all entries of a group with one shared step size, which is proportional to the largest
absolute entry of the group. A group runs either along the key axis, within one column of the state,
or along the value axis, within one row. When a group contains one large entry, its small entries are
rounded with a coarse step, so the error at four bits depends on which entries share a group.
Section~\ref{sec:structure:cols} measures along which axis the large entries of the state cluster,
and Section~\ref{sec:structure:delta} traces this structure to the delta value.

\subsection{Column Structure of the State}
\label{sec:structure:cols}

We compute the state over $2{,}048$-token WikiText-103 sequences on all four backbones. In a representative head the large entries form columns, that is, they occur at the same value index for every key index (Figure~\ref{fig:axis}a). We quantify this with the ratio of the largest absolute entry of a group to the root mean square of its entries (max/RMS). The larger the ratio, the fewer quantization levels a typical entry of the group receives. Groups along
the value axis, which cross columns, have the larger ratio at every group width, and the difference
grows with the width (Figure~\ref{fig:axis}b).

\noindent\textbf{Finding 2.} The large entries of the recurrent state concentrate in columns.

\subsection{Channel Identity of the Delta Value}
\label{sec:structure:delta}
At each token, the recurrence adds the outer product $k_t u_t^\top$ to the decayed state (Eq.~\ref{eq:delta}). A value channel that
is large across many tokens therefore accumulates in one column, and a key coordinate that is large
across many tokens accumulates in one row. Either factor could therefore produce the columns of Section~\ref{sec:structure:cols}. To separate them, we recompute the state from the recurrence inputs with the channels of $u_t$ permuted at
every token, which keeps the values of $u_t$ but assigns them to a different channel at each token, and separately
with the entries of $k_t$ permuted. With $u_t$ permuted, key-axis groups have the larger max/RMS ratio on all four backbones, and with $k_t$ permuted the difference between the two axes widens (Figure~\ref{fig:axis}c).

\noindent\textbf{Finding 3.} The columns of large entries in the state come from channels of the delta
value $u_t$ that stay large across tokens.

\section{Method}
\label{sec:method}
PACE has four components (Figure~\ref{fig:method_overview}): periodic rounding \cnum{A} with a
decode buffer \cnum{B} (Section~\ref{sec:method:cost}), which follow from Section~\ref{sec:period};
channel-aligned scale groups \cnum{C} (Section~\ref{sec:method:quant}), from Findings~2 and~3; and
head-level precision allocation \cnum{D} (Section~\ref{sec:method:heads}), from Finding~1. PACE-4 uses
\cnum{A}--\cnum{C}; PACE-4/8 adds \cnum{D}.

\subsection{Periodic Rounding with a Decode Buffer (\cnum{A}\cnum{B})}
\label{sec:method:cost}
\begin{figure*}[h]
\centering
\includegraphics[width=\textwidth]{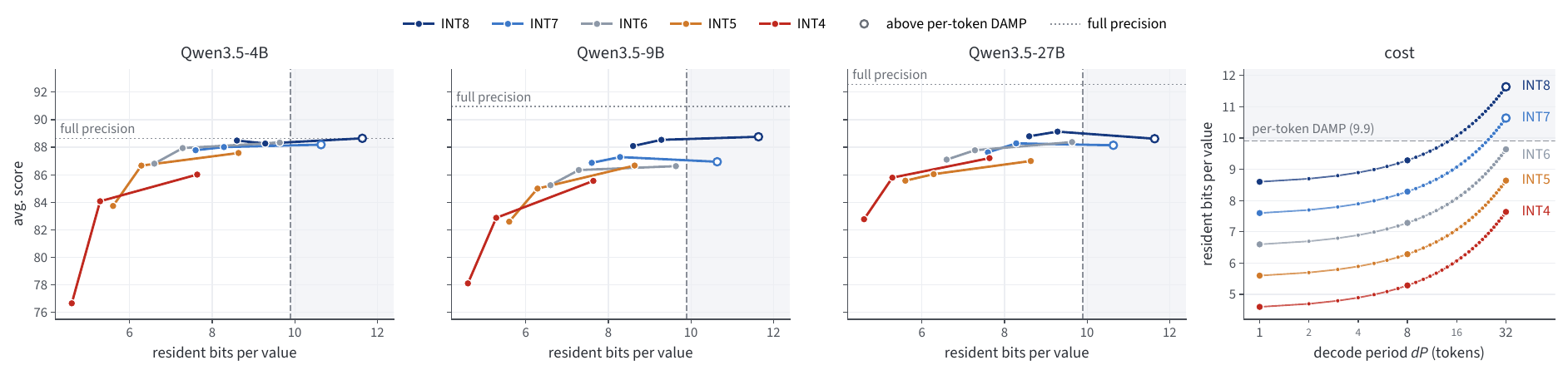}
\caption{Mean score over GSM8K, IFEval and RULER-32k against resident bits per value for uniform
value-axis states of four to eight bits on the three Qwen3.5 backbones (Qwen3.6-35B-A3B in
Appendix~\ref{app:period}). Each line is one width, with $dP = 1, 8, 32$ from left to
right; dotted: full precision; shaded: above the $9.9$ bits of DAMP INT8. Right: resident bits per value
against $dP$ (Eq.~\ref{eq:resident}).}
\label{fig:period}
\end{figure*}

Rounding every $dP$ tokens requires holding the tokens generated since the last rounding in the
decode buffer, which adds to the memory of each configuration. Prior work reports only the stored
state, that is, the $b$-bit payload and its group scales \citep{zhang2026damp}; with one fp16 scale
per $g$ values, this gives $b + 16/g$ bits per value. We define \emph{resident bits per value} as the
bits of the stored state and the buffer together, divided by the number of state values. The buffer holds
$(a_t, k_t, v_t)$ per token, the key once per key head and the value once per value head, so with keys
at $b_k$ and values at $b_v$ bits it adds per state value
\begin{equation}
c\,dP, \qquad c = \frac{b_k\,d_k/\gamma + b_v\,d_v + 32}{d_k d_v} ,
\label{eq:resident}
\end{equation}
where $\gamma$ is the number of value heads sharing a key head. At $dP = 1$ the buffer holds only the current token. Resident bits in Figure~\ref{fig:period} and Appendix~\ref{app:period} include the buffer;
Table~\ref{tab:thru_mem} does not. In uniform configurations, which hold every head at the same bit width, the buffer holds keys and
values at eight bits. In the mixed configuration of PACE-4/8, which holds a quarter of the heads at
eight bits, it holds the keys at four bits and the values in bf16, since
rounding the buffered values below eight bits reduces accuracy more than rounding the keys
(Appendix~\ref{app:keybuf}).

\subsection{Channel-Aligned Scale Groups (\cnum{C})}
\label{sec:method:quant}
We store the state at four bits in channel-aligned scale groups and keep the per-token decay and the
group scales at higher precision.

\paragraph{Scale groups.}
We divide each head's state $S \in \mathbb{R}^{d_k \times d_v}$ into groups of $g$ consecutive entries
along the key axis, so that each group holds $g$ entries of one column, and store one fp16 scale per
group (Figure~\ref{fig:method_overview}, \cnum{C}). Uniform configurations round symmetrically with a step size proportional to the absolute maximum of the group at $b + 16/g$ bits per value; the mixed configuration uses an affine
grid per group whose zero point is stored as a $b$-bit code, at $b + (16+b)/32$ bits per value (Appendix~\ref{app:zp}). Prior work groups along the value axis
\citep{zhang2026damp}. We call our groups channel-aligned because each takes its entries from one
channel of the delta value (Finding~3), whereas a group along the value axis spans columns of different
magnitude, and its step size follows the largest of them. We round to nearest throughout.

\subsection{Head-Level Precision Allocation (\cnum{D})}
\label{sec:method:heads}
We hold the heads that retain rounding error longest at eight bits and the rest at four, applying the allocation criterion of \citet{zhang2026damp} per head instead of per channel
(Finding~1). For each head $h$, we measure the energy $e_h$ of the error that one rounding adds and
assign eight bits to the heads with the largest retained error $E_h$ (Eq.~\ref{eq:retained}). The ranking is computed once per checkpoint on eight
held-out windows of $2{,}048$ tokens. With a quarter of the heads per layer at $8$ bits, the payload is
$0.75 \cdot 4.625 + 0.25 \cdot 8.75 = 5.66$ bits per value, the same as DAMP INT4.

We evaluate two variants: PACE-4, a uniform four-bit state with symmetric key-axis groups of $64$ and
an int8 buffer, and PACE-4/8, which adds head allocation, uses key-axis groups of $32$ with an integer
zero point (Appendix~\ref{app:zp}), and holds the buffered keys at four bits and the values in bf16.

\section{Experimental Setup}
\label{sec:setup}

\paragraph{Models.}
We evaluate Qwen3.5-4B, 9B and 27B and Qwen3.6-35B-A3B \citep{qwen2026qwen35,qwen2026qwen36}, hybrids that
interleave gated-delta layers with full-attention layers. The gated-delta layers run on the \texttt{fla} kernels. 

\paragraph{Benchmarks.}
We evaluate on GSM8K \citep{cobbe2021gsm8k}, MMLU-Pro \citep{wang2024mmlupro} on a subset of $40$ questions per subject ($n{=}560$), IFEval \citep{zhou2023ifeval} ($n{=}541$), RULER variable tracking \citep{hsieh2024ruler} at $32{,}768$ tokens, and MATH-500
\citep{hendrycks2021math,lightman2024verify}. Perplexity is measured on WikiText-103 \citep{merity2017wikitext}.
All generative tasks
are zero-shot with the chat template and thinking disabled.

\paragraph{Baselines.}
We compare against the full-precision state, against DAMP \citep{zhang2026damp} and against the
decoupled scale quantization (DSQ) of Q-Mamba \citep{chen2025qmamba}, which we reimplement. DAMP INT8 follows the published
configuration: value-axis groups of $32$ at eight bits, and the $16$ of $128$ key channels per head with the
highest error--persistence score held in fp16. It rounds the state after every generated token and
serves as our reference for resident memory; we list it at its published $9.9$ bits per value. DAMP INT4 applies the same construction at a four-bit base, with $32$ key channels per head held
at eight bits. Q-Mamba DSQ scales the state by one fp16 vector per axis, computed at each rounding; we
implement the quantizer without the fine-tuning step of the published method.

\paragraph{Protocol.}
All benchmarks run in the LM Evaluation Harness \citep{gao2024lmeval} with greedy decoding. Generation
is capped at $1{,}024$ tokens for GSM8K and MATH-500 and $8{,}192$ for MMLU-Pro. We report GSM8K (flexible-extract filter) and MATH-500 as exact match, IFEval as prompt-level strict
accuracy, and RULER as variable-tracking accuracy. Unless stated otherwise, every quantized configuration uses a four-bit base
width and an int8 decode buffer at $dP{=}8$. Tables~\ref{tab:headline} and~\ref{tab:ablation} and Appendix~\ref{app:period} come from separate evaluation runs, each with its own full-precision row.

\section{Results}
\label{sec:res}
We first isolate the decode period (Section~\ref{sec:res:period}), then compare PACE with DAMP and
Q-Mamba (Section~\ref{sec:headline}), test its
components one at a time (Section~\ref{sec:ablate}), and measure state drift
(Section~\ref{sec:res:drift}) and decoding throughput (Section~\ref{sec:res:thru}).

\subsection{Effect of the Decode Period}\label{sec:res:period}
To isolate the period, we use the quantizer of prior work, uniform round-to-nearest with value-axis
groups of $32$, at four to eight bits and $dP \in \{1, 8, 32\}$ (Figure~\ref{fig:period}; per-benchmark scores in Appendix~\ref{app:period}). Per-token rounding loses more accuracy as the width falls, and rounding every eight tokens scores above
per-token rounding at every width below eight bits on all four backbones. 

The default period of eight closes most of the gap that a period of $32$ closes, with a quarter of
the buffer memory, and matches the flush interval of DeltaLog \citep{lin2026deltalog}. On the three larger backbones a gap to full
precision remains at every width, concentrated in IFEval and RULER-32k. The RULER-32k gap is similar at all three decode periods, which suggests that this part of the gap arises in prefill.

\subsection{Main Results}\label{sec:headline}
Table~\ref{tab:headline} compares PACE with DAMP and Q-Mamba DSQ on four backbones. On the three larger backbones PACE-4/8 scores below DAMP INT8 on
IFEval and close to it on average, and the other benchmarks differ in both directions. On Qwen3.5-4B its
average is lower, with lower scores on GSM8K, MMLU-Pro and IFEval.

\input{tabs/tab_headline}
At the resident bits of DAMP INT4, PACE-4/8 with an int8 buffer scores above DAMP INT4 on every
backbone, with the largest difference on Qwen3.6-35B-A3B. With its mixed buffer, PACE-4/8 scores
higher than this variant on the three larger backbones and lower on Qwen3.5-4B. PACE-4 holds the
same resident bits as Q-Mamba DSQ and scores higher on every backbone.

\subsection{Ablations}\label{sec:ablate}
We compare state quantizers on Qwen3.5-4B and 27B, changing one component at a time, and report
IFEval with a $95\%$ interval on the paired difference to the full-precision state and perplexity on
WikiText-103 at $32{,}768$ tokens (Table~\ref{tab:ablation}).

Groups along the key axis yield a lower perplexity than groups along the value axis at the
same width, with and without head allocation, and a higher IFEval score without it. Q-Mamba DSQ yields a higher perplexity than key-axis
groups at the same resident bits per value.

At equal resident bits, head-level allocation yields a lower perplexity than DAMP INT4.

\input{tabs/tab_ablation}

\subsection{State Drift under Periodic Rounding}\label{sec:res:drift}
\begin{figure}[h]
\centering
\includegraphics[width=\linewidth]{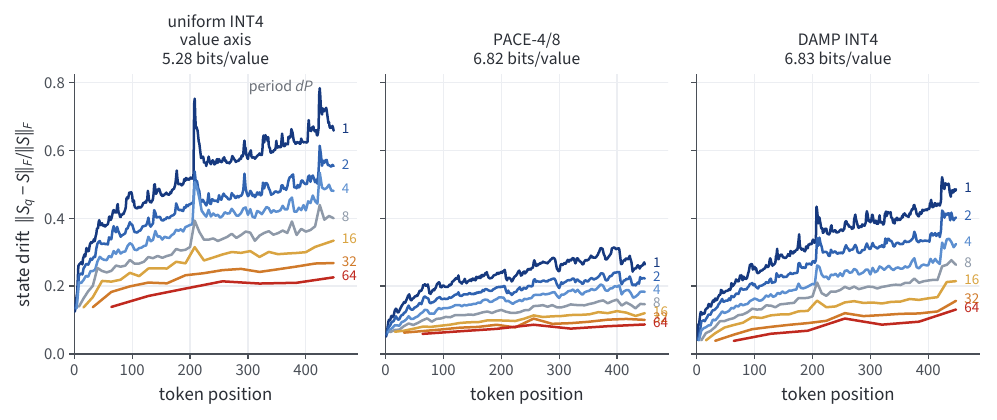}
\caption{Relative state error against token position on Qwen3.5-4B, one curve per period $dP$. The DAMP panel (right) keeps $16$
key channels per head in fp16.}
\label{fig:drift}
\end{figure}

To measure the effect of the rounding period on the state itself, we compare the quantized state with
the full-precision state at the same token. We run Qwen3.5-4B on the first 512 tokens of the
WikiText-103 test set \citep{merity2017wikitext}, round the state every $dP$ tokens over the whole sequence, for $dP$ from 1 to 64, and report the median over recurrent layers of the relative error $\|S_q - S\|_F / \|S\|_F$, which
we call drift. We compare three four-bit
quantizers: uniform INT4 with value-axis groups, DAMP INT4, and the state
quantizer of PACE-4/8 (Figure~\ref{fig:drift}).

For all three quantizers, drift decreases as $dP$ grows at every token position, and the
same ordering holds on Qwen3.6-35B-A3B (Appendix~\ref{app:drift}). Drift grows more
slowly than the number of roundings, which is consistent with a decaying recurrence that removes old
rounding error while each rounding adds new error (Section~\ref{sec:period:error}).

\subsection{Throughput}\label{sec:res:thru}
\begin{figure}[h]
\centering
\includegraphics[width=0.8\linewidth]{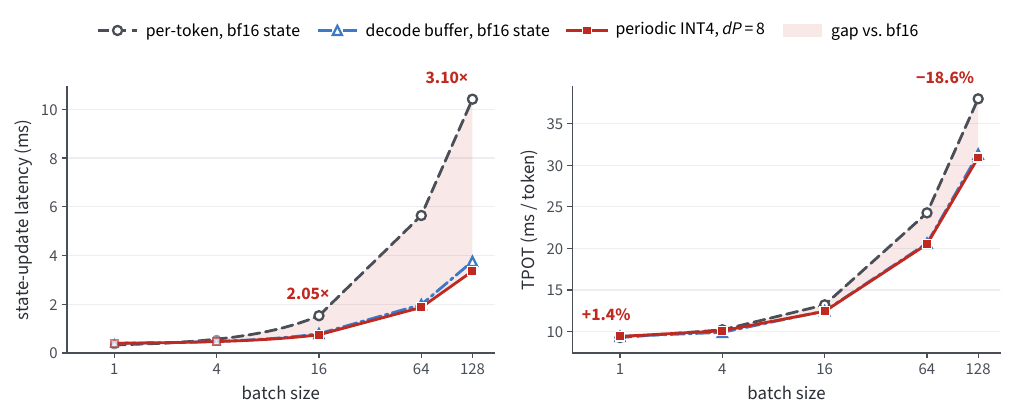}
\caption{Decoding cost on Qwen3.5-4B (H100 PCIe, 512-token context) against batch size. Left:
latency of the recurrent-state update. Right: time per output token. The shading marks the gap to bf16. Labels give the ratio to bf16 (left) and the relative change in
TPOT (right) at the ends of each curve.}
\label{fig:throughput}
\end{figure}

We measure decoding cost on Qwen3.5-4B on one H100 PCIe GPU at a $512$-token context and batch sizes
from $1$ to $128$. We compare per-token decoding with a bf16 state, the decode buffer with a bf16 state
\citep{zou2026kvbuffer}, and periodic decoding at $dP{=}8$ with a uniform INT4 state and an int8
buffer, for which we implement Triton kernels that apply the buffered tokens to the packed state. We
report time per output token (TPOT) and the latency of the state update (Figure~\ref{fig:throughput}, Appendix~\ref{app:mem}).

Periodic rounding lets a four-bit state keep the speedup of the decode buffer. Periodic INT4 decoding
has a lower TPOT than per-token bf16 decoding from batch 16 upward, and the difference grows with batch
size, as reading and writing the per-sequence state take a growing share of each decoding step. The decode
buffer with a bf16 state accounts for most of this reduction, and the INT4 state reduces TPOT slightly
further.
The memory each configuration holds per sequence, measured in the same run, agrees with the resident bits of Eq.~\ref{eq:resident} (Appendix~\ref{app:mem}). 

\section{Conclusion}
\label{sec:conc}
We introduced PACE, a post-hoc four-bit state quantizer for gated-delta models that rounds the state
once every few generated tokens and holds the tokens in between in a decode buffer. For uniform
configurations, a period of eight tokens reduces the accuracy loss at every width below eight bits on
four backbones. This is consistent with the bound of Section~\ref{sec:period}, under which more decay
between two roundings leaves less of each rounding error in the state. The large entries of the state
gather in columns that follow the channels of the delta value, and PACE forms each scale group from
the entries of one such column. PACE-4/8 comes close to the accuracy of DAMP INT8 on the three larger
backbones at about two thirds of its resident memory, and PACE-4 scores higher than Q-Mamba DSQ at
equal resident bits. Periodic rounding also keeps the decoding speedup of the decode buffer.

\section*{Limitations}
\label{sec:lim}
We evaluate four hybrid models from the Qwen3.5 and Qwen3.6 families, all of which use the gated delta rule.
Periodic rounding applies to any linear-attention layer whose state can be advanced over several tokens
at once, which includes the chunked forms of gated linear attention, Mamba-2 and DeltaNet variants.
The head decays that enter the error bound, the column structure that motivates key-axis scale groups,
and the head ranking are measured only on these models. None of them is shown to carry over to
other model families, to models whose layers are all recurrent, or to token mixers with a different
update or decay.

Periodic rounding changes when the state is rounded and leaves the choice of quantizer open. Besides PACE, we evaluate DAMP INT4 and Q-Mamba DSQ under the same period. We do not test quantizers
with static scales, stochastic rounding or rotations of the state under periodic rounding.

The configuration space of periodic state quantization is large. It spans the prefill and decode
periods, the bit widths of the state and of the buffered keys and values, the axis and width of the
scale groups, and the fraction of heads held at eight bits, and we explore only part of it. The
prefill period in particular is fixed at the $256$-token chunk of the kernel in all benchmark evaluations, so
the effect of rounding the state more or less often during prefill is not measured.
Throughput is timed only for a uniform four-bit state; the key-axis groups and head allocation of PACE
are not timed.

\section*{Acknowledgments}
Alan Akbik and Patrick Haller are
supported by the Deutsche Forschungsgemeinschaft (DFG, German Research Foundation) under Emmy Noether grant ``Eidetic Representations of Natural Language'' (project number
448414230). Alan Akbik is furthermore supported under Germany's Excellence Strategy ``Science of Intelligence'' (EXC 2002/1, project number
390523135).
The authors gratefully acknowledge the scientific support and HPC resources provided by the Erlangen National High Performance Computing Center (NHR@FAU) of the Friedrich-Alexander-Universit\"at Erlangen-N\"urnberg (FAU) under the NHR project ``Pre-Training SLMs via Knowledge Distillation''. NHR funding is provided by federal and Bavarian state authorities.

\bibliography{custom}

\clearpage
\appendix
% Table 3, set below the heading of Appendix A.
\newcommand{\openingblock}{{%
\centering
\begingroup
\renewenvironment{table*}[1][]{}{}%
\renewcommand{\caption}[1]{}%
\renewcommand{\label}[1]{}%
\begin{minipage}[t]{0.49\textwidth}\centering{\small (a) Qwen3.5-4B}\\[2pt]\resizebox{\linewidth}{!}{\input{tabs/tab_opening_4b}}\end{minipage}\hfill
\begin{minipage}[t]{0.49\textwidth}\centering{\small (b) Qwen3.5-9B}\\[2pt]\resizebox{\linewidth}{!}{\input{tabs/tab_opening_9b}}\end{minipage}\\[8pt]
\begin{minipage}[t]{0.49\textwidth}\centering{\small (c) Qwen3.5-27B}\\[2pt]\resizebox{\linewidth}{!}{\input{tabs/tab_opening_27b}}\end{minipage}\hfill
\begin{minipage}[t]{0.49\textwidth}\centering{\small (d) Qwen3.6-35B-A3B}\\[2pt]\resizebox{\linewidth}{!}{\input{tabs/tab_opening_35b}}\end{minipage}
\endgroup
\captionof{table}{Results on post-hoc state quantization on four backbones: (a) Qwen3.5-4B, (b) Qwen3.5-9B, (c) Qwen3.5-27B and (d) Qwen3.6-35B-A3B. Every row rounds the state to nearest in value-axis groups of $32$. The prefill period is $P{=}256$ throughout, so $dP$ changes only decoding. \textbf{resident} gives resident bits per value (Eq.~\ref{eq:resident}), including the decode buffer at $0.098$ bits per value per token of the period. \textbf{avg} is the unweighted mean of GSM8K, IFEval and RULER-$32$k.}
\label{tab:opening}
\vspace{1.5em}}}
% One column: appendix floats stay in their own sections, so no heading is left empty.
\renewenvironment{figure}[1][]{\origfigure[H]}{\endorigfigure}
\renewenvironment{table}[1][]{\origtable[H]}{\endorigtable}
\renewenvironment{figure*}[1][]{\begin{figure}}{\end{figure}}
\renewenvironment{table*}[1][]{\begin{table}}{\end{table}}

\section{Per-Benchmark Scores across Widths and Periods}
\label{app:period}
Table~\ref{tab:opening} gives the per-benchmark scores behind Figure~\ref{fig:period}.
\begin{table}[H]{%
\centering
\begingroup
\renewenvironment{table*}[1][]{}{}%
\renewcommand{\caption}[1]{}%
\renewcommand{\label}[1]{}%
\begin{minipage}[t]{0.49\textwidth}\centering{\small (a) Qwen3.5-4B}\\[2pt]\resizebox{\linewidth}{!}{\input{tabs/tab_opening_4b}}\end{minipage}\hfill
\begin{minipage}[t]{0.49\textwidth}\centering{\small (b) Qwen3.5-9B}\\[2pt]\resizebox{\linewidth}{!}{\input{tabs/tab_opening_9b}}\end{minipage}\\[8pt]
\begin{minipage}[t]{0.49\textwidth}\centering{\small (c) Qwen3.5-27B}\\[2pt]\resizebox{\linewidth}{!}{\input{tabs/tab_opening_27b}}\end{minipage}\hfill
\begin{minipage}[t]{0.49\textwidth}\centering{\small (d) Qwen3.6-35B-A3B}\\[2pt]\resizebox{\linewidth}{!}{\input{tabs/tab_opening_35b}}\end{minipage}
\endgroup
\captionof{table}{Results on post-hoc state quantization on four backbones: (a) Qwen3.5-4B, (b) Qwen3.5-9B, (c) Qwen3.5-27B and (d) Qwen3.6-35B-A3B. Every row rounds the state to nearest in value-axis groups of $32$. The prefill period is $P{=}256$ throughout, so $dP$ changes only decoding. \textbf{resident} gives resident bits per value (Eq.~\ref{eq:resident}), including the decode buffer at $0.098$ bits per value per token of the period. \textbf{avg} is the unweighted mean of GSM8K, IFEval and RULER-$32$k.}
\label{tab:opening}
\vspace{1.5em}}\end{table}

\section{State Drift}
\label{app:drift}

\begin{figure}[h]
\includegraphics[width=\linewidth]{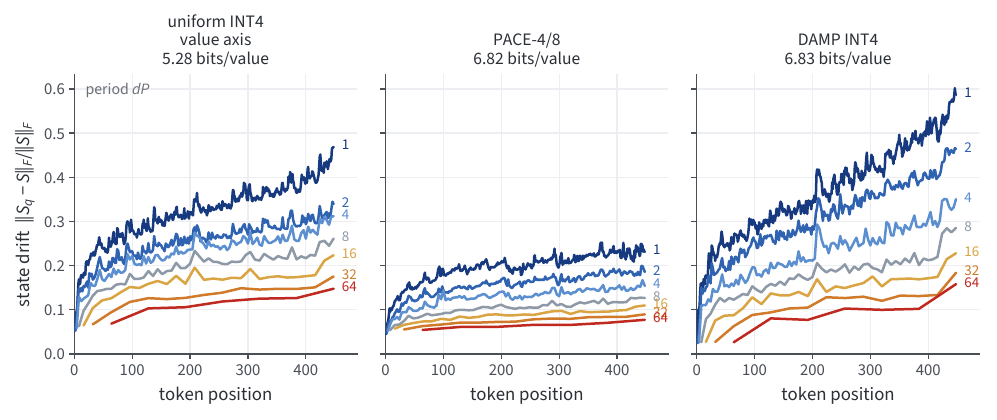}
\caption{State drift on Qwen3.6-35B-A3B, measured as in Figure~\ref{fig:drift}.}
\label{fig:drift:35b}
\end{figure}

\section{State Statistics along the Two Axes}\label{app:axis}
A two-way additive decomposition of $\log|S|$ within each head assigns more variance to the row
(key-index) effect than to the column effect on all four backbones, so this bulk statistic ranks the
two axes opposite to the max/RMS ratio. The step size of a group follows its largest entry, which the tail statistics in
Figure~\ref{fig:axis} and Table~\ref{tab:permute} measure.
\input{tabs/tab_permute}

\section{Separate Key and Value Buffer Widths}\label{app:keybuf}
To locate which buffered tensor limits the buffer width, we round only the keys or only the values to
four bits and keep the other in bf16, with the state quantizer of PACE-4/8 at $dP{=}8$. Rounding the
values reduces accuracy more than rounding the keys (Table~\ref{tab:keybuf}). The rank-one form of the
update predicts the opposite order, since a key error misaddresses both the read and the write, while a
value error affects only the write. Rounding either tensor to four
bits gives a similar relative input error, but the kernel L2-normalizes the buffered keys after
rounding, so only the direction of a key error reaches the state, while a value error enters it
unchanged.

\input{tabs/tab_keybuf}

\section{Integer Zero Point}\label{app:zp}
The affine grid needs a zero point, and storing it as a float means an entry that has decayed to zero
is rounded to a nonzero value at every rounding. We snap the zero point onto the payload grid and store it
as a $b$-bit code, so the side information is $(16 + b)/32$ bits per value and a decayed entry rounds
to zero by construction. An affine $4$-bit group takes $4.625$ bits per value and an $8$-bit group
$8.75$.

\section{Throughput Measurement and Memory}
\label{app:mem}
The state-update latency is the TPOT minus the TPOT of the same model with the recurrent operator
replaced by a constant. All timings use CUDA graphs and the median over $15$ timing blocks.
% Source: /vol/tmp2/hallepat/e2e_scratch/b179_nq_m8.json (B179, Qwen3.5-4B, H100 PCIe), bytes_per_seq,
% identical at every batch size. Resident bits from Eq. (resident): INT4 g32 payload 4.50 + 8 x 0.098;
% the BF16 buffer arm stores unit scales, 16.50 + 8 x 0.098.
\begin{table}[h]
\centering
\resizebox{\colfigwidth}{!}{
\centering\small
\setlength{\tabcolsep}{4pt}
\begin{tabular}{@{}lrrrrr@{}}
\toprule
\textbf{configuration} & \textbf{state} & \textbf{buffer} & \textbf{other} & \textbf{bits/value} & \textbf{Eq.~\ref{eq:resident}} \\
\midrule
per-token, bf16 state & $24.00$ & \textendash & \textendash & $16.00$ & $16.00$ \\
decode buffer, bf16 state & $24.75$ & $1.14$ & $0.08$ & $17.32$ & $17.28$ \\
periodic INT4, $dP{=}8$ & $6.75$ & $1.14$ & $0.08$ & $5.32$ & $5.28$ \\
\bottomrule
\end{tabular}
}
\caption{Recurrent-state memory per sequence on Qwen3.5-4B in MiB, measured in the run of
Figure~\ref{fig:throughput}, and the bits per state value it implies, next to the resident bits of
Eq.~\ref{eq:resident}. \textbf{other} holds the buffered decays and kernel scratch space.}
\label{tab:thru_mem}
\end{table}

\end{document}

%% file: tabs/tab_headline.tex
% generated by tabs/gen_headline.py -- do not edit by hand
\begin{table*}[t]
\centering\small
\setlength{\tabcolsep}{4.5pt}
\begin{adjustbox}{max width=\textwidth}
\begin{tabular}{@{}lrrrrrrr@{}}
\toprule
\textbf{state} & \textbf{bits/value} & \textbf{GSM8K} & \textbf{MMLU-Pro} & \textbf{IFEval} & \textbf{RULER-$32$k} & \textbf{MATH-500} & \textbf{avg} \\
\midrule
\multicolumn{8}{@{}l}{\emph{Qwen3.5\,4B}} \\
\quad full precision state & \cellcolor{cost!70}$32.00$ & 89.2 & 72.3 & 83.3 & 100.0 & 59.0 & 80.8 \\
\quad DAMP INT8, per token & \cellcolor{cost!49}$9.90$ & 89.1 & 73.4 & 82.8 & 100.0 & 59.0 & 80.9 \\
\quad DAMP INT4 & \cellcolor{cost!20}$6.44$ & 89.0 & 71.8 & 80.6 & 100.0 & 58.2 & 79.9 \\
\quad Q-Mamba DSQ & \cellcolor{cost!9}$5.03$ & 88.0 & 63.2 & 74.9 & 97.3 & 52.4 & 75.2 \\
\quad PACE-4/8, int8 buffer & \cellcolor{cost!20}$6.44$ & 88.7 & 71.1 & 81.5 & 100.0 & 62.6 & 80.8 \\
\rowcolor{good!12}\quad \textbf{PACE-4/8} & \cellcolor{cost!23}$6.82$ & \textbf{86.5} & \textbf{70.4} & \textbf{78.2} & \textbf{100.0} & \textbf{60.4} & \textbf{79.1} \\
\rowcolor{good!12}\quad \textbf{PACE-4} & \cellcolor{cost!9}$5.03$ & \textbf{88.6} & \textbf{71.2} & \textbf{80.6} & \textbf{99.3} & \textbf{58.4} & \textbf{79.6} \\
\addlinespace[3pt]
\multicolumn{8}{@{}l}{\emph{Qwen3.5\,9B}} \\
\quad full precision state & \cellcolor{cost!70}$32.00$ & 94.2 & 74.8 & 84.5 & 100.0 & 68.8 & 84.5 \\
\quad DAMP INT8, per token & \cellcolor{cost!49}$9.90$ & 94.1 & 71.8 & 83.4 & 100.0 & 69.4 & 83.7 \\
\quad DAMP INT4 & \cellcolor{cost!20}$6.44$ & 91.5 & 70.5 & 78.0 & 93.3 & 65.2 & 79.7 \\
\quad Q-Mamba DSQ & \cellcolor{cost!9}$5.03$ & 90.4 & 63.6 & 70.6 & 88.0 & 51.6 & 72.8 \\
\quad PACE-4/8, int8 buffer & \cellcolor{cost!20}$6.44$ & 91.3 & 70.4 & 78.6 & 98.7 & 68.0 & 81.4 \\
\rowcolor{good!12}\quad \textbf{PACE-4/8} & \cellcolor{cost!23}$6.82$ & \textbf{92.2} & \textbf{77.0} & \textbf{81.3} & \textbf{100.0} & \textbf{68.0} & \textbf{83.7} \\
\rowcolor{good!12}\quad \textbf{PACE-4} & \cellcolor{cost!9}$5.03$ & \textbf{91.0} & \textbf{70.2} & \textbf{76.5} & \textbf{98.3} & \textbf{63.8} & \textbf{80.0} \\
\addlinespace[3pt]
\multicolumn{8}{@{}l}{\emph{Qwen3.5\,27B}} \\
\quad full precision state & \cellcolor{cost!70}$32.00$ & 96.4 & 83.8 & 86.9 & 100.0 & 70.2 & 87.5 \\
\quad DAMP INT8, per token & \cellcolor{cost!49}$9.90$ & 96.4 & 83.8 & 88.0 & 100.0 & 69.0 & 87.4 \\
\quad DAMP INT4 & \cellcolor{cost!20}$6.44$ & 96.1 & 81.1 & 79.1 & 92.7 & 70.2 & 83.8 \\
\quad Q-Mamba DSQ & \cellcolor{cost!9}$5.03$ & 96.0 & 78.0 & 73.2 & 87.3 & 59.8 & 78.9 \\
\quad PACE-4/8, int8 buffer & \cellcolor{cost!20}$6.44$ & 96.4 & 81.6 & 79.9 & 95.0 & 69.8 & 84.5 \\
\rowcolor{good!12}\quad \textbf{PACE-4/8} & \cellcolor{cost!23}$6.82$ & \textbf{96.5} & \textbf{82.5} & \textbf{86.1} & \textbf{100.0} & \textbf{71.6} & \textbf{87.3} \\
\rowcolor{good!12}\quad \textbf{PACE-4} & \cellcolor{cost!9}$5.03$ & \textbf{95.9} & \textbf{81.2} & \textbf{79.3} & \textbf{96.0} & \textbf{68.8} & \textbf{84.2} \\
\addlinespace[3pt]
\multicolumn{8}{@{}l}{\emph{Qwen3.6\,35B-A3B}} \\
\quad full precision state & \cellcolor{cost!70}$32.00$ & 96.3 & 82.1 & 83.7 & 100.0 & 70.6 & 86.5 \\
\quad DAMP INT8, per token & \cellcolor{cost!49}$9.90$ & 96.1 & 83.4 & 83.0 & 100.0 & 69.2 & 86.3 \\
\quad DAMP INT4 & \cellcolor{cost!20}$6.44$ & 93.9 & 66.6 & 69.9 & 71.0 & 43.8 & 69.0 \\
\quad Q-Mamba DSQ & \cellcolor{cost!9}$5.03$ & 93.6 & 72.5 & 63.2 & 66.7 & 59.8 & 71.2 \\
\quad PACE-4/8, int8 buffer & \cellcolor{cost!20}$6.44$ & 95.5 & 77.7 & 74.1 & 84.3 & 60.0 & 78.3 \\
\rowcolor{good!12}\quad \textbf{PACE-4/8} & \cellcolor{cost!23}$6.82$ & \textbf{95.3} & \textbf{80.5} & \textbf{81.0} & \textbf{100.0} & \textbf{68.4} & \textbf{85.0} \\
\rowcolor{good!12}\quad \textbf{PACE-4} & \cellcolor{cost!9}$5.03$ & \textbf{95.0} & \textbf{77.0} & \textbf{72.6} & \textbf{83.0} & \textbf{58.8} & \textbf{77.3} \\
\bottomrule
\end{tabular}
\end{adjustbox}
\caption{PACE-4/8 and PACE-4 (shaded) hold $0.69$ and $0.51$ of the resident bits of DAMP INT8. Bits per value are resident bits (Eq.~\ref{eq:resident}); all configurations except DAMP INT8 round the state every eight generated tokens ($dP{=}8$). The average is the mean over all benchmarks.}
\label{tab:headline}
\end{table*}

%% file: tabs/tab_ablation.tex
% generated by tabs/gen_ablation.py -- do not edit by hand
\begin{table*}[t]
\centering\small
\setlength{\tabcolsep}{6pt}
\begin{tabular}{@{}lrrrr@{}}
\toprule
\textbf{configuration} & \textbf{bits} & \textbf{IFEval 4B} & \textbf{IFEval 27B} & \textbf{ppl 4B} \\
\midrule
full precision & $32.00$ & $82.26$ & $86.69$ & $7.713$ \\
\midrule
\multicolumn{5}{@{}l}{\emph{Scale structure}} \\
\quad value axis, $g{=}32$ & $5.28$ & $73.75_{\pm3.29}$ & $75.05_{\pm3.24}$ & $8.158$ \\
\quad Q-Mamba DSQ \citep{chen2025qmamba} & $5.03$ & $74.86_{\pm3.22}$ & $73.94_{\pm3.33}$ & $8.320$ \\
\quad key axis, $g{=}32$ & $5.28$ & $80.59_{\pm2.53}$ & $78.37_{\pm2.84}$ & $7.719$ \\
\quad key axis, $g{=}64$ & $5.03$ & $80.59_{\pm2.73}$ & $78.93_{\pm2.78}$ & $7.734$ \\
\addlinespace[3pt]
\multicolumn{5}{@{}l}{\emph{Precision allocation}} \\
\quad channel protection, value axis (DAMP INT4) & $6.44$ & $79.48_{\pm2.68}$ & $\mathbf{79.85}_{\pm2.91}$ & $8.003$ \\
\quad head allocation, value axis & $6.44$ & $80.41_{\pm2.76}$ & $77.63_{\pm2.96}$ & $7.776$ \\
\quad head allocation, key axis & $6.44$ & $\mathbf{81.52}_{\pm2.51}$ & $78.00_{\pm2.88}$ & $\mathbf{7.687}$ \\
\bottomrule
\end{tabular}
\caption{Scaling the four-bit state along the key axis and allocating eight bits per head reduce the loss of the uniform value-axis quantizer. Bits are resident bits per value at $dP{=}8$ with an int8 decode buffer.}
\label{tab:ablation}
\end{table*}

%% file: tabs/tab_opening_4b.tex
% generated by tabs/gen_opening.py -- do not edit by hand
\begin{table*}[t]
\centering\small
\setlength{\tabcolsep}{6pt}
\begin{tabular}{@{}lr|rrr|r@{}}
\toprule
\textbf{state} & \textbf{resident} & \textbf{GSM8K} & \textbf{IFEval} & \textbf{RULER-$32$k} & \textbf{avg} \\
\midrule
full precision & \cellcolor{cost!72}$32.00$ & \cellcolor{good!50}82.7 & \cellcolor{good!50}83.3 & \cellcolor{good!50}100.0 & \cellcolor{good!50}88.7 \\
INT8, $dP{=}1$ & \cellcolor{cost!39}$8.60$ & \cellcolor{good!50}84.3 & \cellcolor{good!42}81.1 & \cellcolor{good!50}100.0 & \cellcolor{good!49}88.5 \\
INT8, $dP{=}8$ & \cellcolor{cost!44}$9.28$ & \cellcolor{good!50}84.8 & \cellcolor{good!38}80.0 & \cellcolor{good!50}100.0 & \cellcolor{good!48}88.3 \\
INT8, $dP{=}32$ & \cellcolor{cost!61}$11.64$ & \cellcolor{good!50}84.2 & \cellcolor{good!44}81.7 & \cellcolor{good!50}100.0 & \cellcolor{good!50}88.6 \\
\addlinespace[2pt]
INT7, $dP{=}1$ & \cellcolor{cost!31}$7.60$ & \cellcolor{good!50}84.2 & \cellcolor{good!35}79.1 & \cellcolor{good!50}100.0 & \cellcolor{good!47}87.8 \\
INT7, $dP{=}8$ & \cellcolor{cost!36}$8.28$ & \cellcolor{good!50}83.6 & \cellcolor{good!39}80.4 & \cellcolor{good!50}100.0 & \cellcolor{good!47}88.0 \\
INT7, $dP{=}32$ & \cellcolor{cost!54}$10.64$ & \cellcolor{good!50}83.8 & \cellcolor{good!40}80.8 & \cellcolor{good!50}100.0 & \cellcolor{good!48}88.2 \\
\addlinespace[2pt]
INT6, $dP{=}1$ & \cellcolor{cost!24}$6.60$ & \cellcolor{good!49}82.4 & \cellcolor{good!31}78.0 & \cellcolor{good!50}100.0 & \cellcolor{good!43}86.8 \\
INT6, $dP{=}8$ & \cellcolor{cost!29}$7.28$ & \cellcolor{good!50}83.5 & \cellcolor{good!40}80.6 & \cellcolor{good!49}99.7 & \cellcolor{good!47}87.9 \\
INT6, $dP{=}32$ & \cellcolor{cost!46}$9.64$ & \cellcolor{good!50}84.0 & \cellcolor{good!42}81.3 & \cellcolor{good!49}99.7 & \cellcolor{good!49}88.3 \\
\addlinespace[2pt]
INT5, $dP{=}1$ & \cellcolor{cost!16}$5.60$ & \cellcolor{good!42}80.6 & \cellcolor{good!12}70.6 & \cellcolor{good!50}100.0 & \cellcolor{good!32}83.7 \\
INT5, $dP{=}8$ & \cellcolor{cost!21}$6.28$ & \cellcolor{good!50}83.2 & \cellcolor{good!28}77.1 & \cellcolor{good!49}99.7 & \cellcolor{good!42}86.7 \\
INT5, $dP{=}32$ & \cellcolor{cost!39}$8.64$ & \cellcolor{good!50}83.8 & \cellcolor{good!35}79.3 & \cellcolor{good!49}99.7 & \cellcolor{good!46}87.6 \\
\addlinespace[2pt]
INT4, $dP{=}1$ & \cellcolor{cost!9}$4.60$ & \cellcolor{good!11}69.3 & \cellcolor{good!1}61.4 & \cellcolor{good!47}99.3 & \cellcolor{good!14}76.7 \\
INT4, $dP{=}8$ & \cellcolor{cost!14}$5.28$ & \cellcolor{good!41}80.2 & \cellcolor{good!17}73.0 & \cellcolor{good!46}99.0 & \cellcolor{good!33}84.1 \\
INT4, $dP{=}32$ & \cellcolor{cost!31}$7.64$ & \cellcolor{good!47}82.0 & \cellcolor{good!28}77.1 & \cellcolor{good!46}99.0 & \cellcolor{good!40}86.0 \\
\bottomrule
\end{tabular}
\caption{Results on post-training state quantization on Qwen3.5-4B. Every row rounds the state to nearest with a group-$32$ scale. The prompt is rounded once per $256$-token kernel chunk throughout, so $dP$ varies the generation alone. \textbf{resident} is the effective bits stored for state during inference: the payload plus the decode buffer, at $0.098$ bits per value per token of period. \textbf{avg} is the unweighted mean of GSM8K, IFEval and RULER-VT; MMLU-Pro is not in it.}
\label{tab:opening:4b}
\end{table*}

%% file: tabs/tab_opening_9b.tex
% generated by tabs/gen_opening.py -- do not edit by hand
\begin{table*}[t]
\centering\small
\setlength{\tabcolsep}{6pt}
\begin{tabular}{@{}lr|rrr|r@{}}
\toprule
\textbf{state} & \textbf{resident} & \textbf{GSM8K} & \textbf{IFEval} & \textbf{RULER-$32$k} & \textbf{avg} \\
\midrule
full precision & \cellcolor{cost!72}$32.00$ & \cellcolor{good!50}88.4 & \cellcolor{good!50}84.5 & \cellcolor{good!50}100.0 & \cellcolor{good!50}91.0 \\
INT8, $dP{=}1$ & \cellcolor{cost!39}$8.60$ & \cellcolor{good!50}88.4 & \cellcolor{good!28}78.2 & \cellcolor{good!41}97.7 & \cellcolor{good!39}88.1 \\
INT8, $dP{=}8$ & \cellcolor{cost!44}$9.28$ & \cellcolor{good!50}89.0 & \cellcolor{good!30}78.9 & \cellcolor{good!41}97.7 & \cellcolor{good!41}88.5 \\
INT8, $dP{=}32$ & \cellcolor{cost!61}$11.64$ & \cellcolor{good!50}89.2 & \cellcolor{good!31}79.1 & \cellcolor{good!42}98.0 & \cellcolor{good!42}88.8 \\
\addlinespace[2pt]
INT7, $dP{=}1$ & \cellcolor{cost!31}$7.60$ & \cellcolor{good!50}88.5 & \cellcolor{good!21}75.8 & \cellcolor{good!36}96.3 & \cellcolor{good!35}86.9 \\
INT7, $dP{=}8$ & \cellcolor{cost!36}$8.28$ & \cellcolor{good!50}88.6 & \cellcolor{good!26}77.6 & \cellcolor{good!34}95.7 & \cellcolor{good!36}87.3 \\
INT7, $dP{=}32$ & \cellcolor{cost!54}$10.64$ & \cellcolor{good!50}88.9 & \cellcolor{good!22}76.0 & \cellcolor{good!35}96.0 & \cellcolor{good!35}86.9 \\
\addlinespace[2pt]
INT6, $dP{=}1$ & \cellcolor{cost!24}$6.60$ & \cellcolor{good!46}87.3 & \cellcolor{good!15}73.4 & \cellcolor{good!32}95.0 & \cellcolor{good!30}85.2 \\
INT6, $dP{=}8$ & \cellcolor{cost!29}$7.28$ & \cellcolor{good!47}87.7 & \cellcolor{good!22}76.0 & \cellcolor{good!33}95.3 & \cellcolor{good!33}86.3 \\
INT6, $dP{=}32$ & \cellcolor{cost!46}$9.64$ & \cellcolor{good!47}87.6 & \cellcolor{good!24}76.9 & \cellcolor{good!33}95.3 & \cellcolor{good!34}86.6 \\
\addlinespace[2pt]
INT5, $dP{=}1$ & \cellcolor{cost!16}$5.60$ & \cellcolor{good!41}86.1 & \cellcolor{good!6}68.4 & \cellcolor{good!27}93.3 & \cellcolor{good!22}82.6 \\
INT5, $dP{=}8$ & \cellcolor{cost!21}$6.28$ & \cellcolor{good!46}87.5 & \cellcolor{good!18}74.5 & \cellcolor{good!26}93.0 & \cellcolor{good!29}85.0 \\
INT5, $dP{=}32$ & \cellcolor{cost!39}$8.64$ & \cellcolor{good!50}88.6 & \cellcolor{good!26}77.4 & \cellcolor{good!29}94.0 & \cellcolor{good!34}86.7 \\
\addlinespace[2pt]
INT4, $dP{=}1$ & \cellcolor{cost!9}$4.60$ & \cellcolor{good!30}82.6 & \cellcolor{good!0}61.4 & \cellcolor{good!19}90.3 & \cellcolor{good!12}78.1 \\
INT4, $dP{=}8$ & \cellcolor{cost!14}$5.28$ & \cellcolor{good!46}87.3 & \cellcolor{good!11}71.3 & \cellcolor{good!18}90.0 & \cellcolor{good!23}82.9 \\
INT4, $dP{=}32$ & \cellcolor{cost!31}$7.64$ & \cellcolor{good!49}88.1 & \cellcolor{good!24}76.9 & \cellcolor{good!22}91.7 & \cellcolor{good!31}85.6 \\
\bottomrule
\end{tabular}
\caption{Results on post-training state quantization on Qwen3.5-9B. Every row rounds the state to nearest with a group-$32$ scale. The prompt is rounded once per $256$-token kernel chunk throughout, so $dP$ varies the generation alone. \textbf{resident} is the effective bits stored for state during inference: the payload plus the decode buffer, at $0.098$ bits per value per token of period. \textbf{avg} is the unweighted mean of GSM8K, IFEval and RULER-VT; MMLU-Pro is not in it. Widths $4$ and $8$ were measured on NHR G\"ottingen and the widths between them here, both at batch $32$, which is the batch every row of this table shares.}
\label{tab:opening:9b}
\end{table*}

%% file: tabs/tab_opening_27b.tex
% generated by tabs/gen_opening.py -- do not edit by hand
\begin{table*}[t]
\centering\small
\setlength{\tabcolsep}{6pt}
\begin{tabular}{@{}lr|rrr|r@{}}
\toprule
\textbf{state} & \textbf{resident} & \textbf{GSM8K} & \textbf{IFEval} & \textbf{RULER-$32$k} & \textbf{avg} \\
\midrule
full precision & \cellcolor{cost!72}$32.00$ & \cellcolor{good!50}90.8 & \cellcolor{good!50}86.9 & \cellcolor{good!50}100.0 & \cellcolor{good!50}92.6 \\
INT8, $dP{=}1$ & \cellcolor{cost!39}$8.60$ & \cellcolor{good!50}90.8 & \cellcolor{good!27}80.2 & \cellcolor{good!33}95.3 & \cellcolor{good!36}88.8 \\
INT8, $dP{=}8$ & \cellcolor{cost!44}$9.28$ & \cellcolor{good!50}91.3 & \cellcolor{good!29}80.8 & \cellcolor{good!33}95.3 & \cellcolor{good!37}89.1 \\
INT8, $dP{=}32$ & \cellcolor{cost!61}$11.64$ & \cellcolor{good!50}91.6 & \cellcolor{good!23}78.9 & \cellcolor{good!33}95.3 & \cellcolor{good!35}88.6 \\
\addlinespace[2pt]
INT7, $dP{=}1$ & \cellcolor{cost!31}$7.60$ & \cellcolor{good!50}91.1 & \cellcolor{good!20}77.8 & \cellcolor{good!29}94.0 & \cellcolor{good!32}87.6 \\
INT7, $dP{=}8$ & \cellcolor{cost!36}$8.28$ & \cellcolor{good!50}91.2 & \cellcolor{good!24}79.3 & \cellcolor{good!30}94.3 & \cellcolor{good!34}88.3 \\
INT7, $dP{=}32$ & \cellcolor{cost!54}$10.64$ & \cellcolor{good!49}90.6 & \cellcolor{good!25}79.5 & \cellcolor{good!30}94.3 & \cellcolor{good!34}88.1 \\
\addlinespace[2pt]
INT6, $dP{=}1$ & \cellcolor{cost!24}$6.60$ & \cellcolor{good!50}91.5 & \cellcolor{good!15}75.8 & \cellcolor{good!29}94.0 & \cellcolor{good!31}87.1 \\
INT6, $dP{=}8$ & \cellcolor{cost!29}$7.28$ & \cellcolor{good!50}91.8 & \cellcolor{good!17}76.5 & \cellcolor{good!32}95.0 & \cellcolor{good!33}87.8 \\
INT6, $dP{=}32$ & \cellcolor{cost!46}$9.64$ & \cellcolor{good!50}92.0 & \cellcolor{good!25}79.5 & \cellcolor{good!28}93.7 & \cellcolor{good!35}88.4 \\
\addlinespace[2pt]
INT5, $dP{=}1$ & \cellcolor{cost!16}$5.60$ & \cellcolor{good!50}91.4 & \cellcolor{good!13}74.7 & \cellcolor{good!20}90.7 & \cellcolor{good!26}85.6 \\
INT5, $dP{=}8$ & \cellcolor{cost!21}$6.28$ & \cellcolor{good!50}90.8 & \cellcolor{good!18}76.7 & \cellcolor{good!20}90.7 & \cellcolor{good!27}86.0 \\
INT5, $dP{=}32$ & \cellcolor{cost!39}$8.64$ & \cellcolor{good!50}91.4 & \cellcolor{good!22}78.6 & \cellcolor{good!20}91.0 & \cellcolor{good!30}87.0 \\
\addlinespace[2pt]
INT4, $dP{=}1$ & \cellcolor{cost!9}$4.60$ & \cellcolor{good!41}88.5 & \cellcolor{good!5}69.5 & \cellcolor{good!19}90.3 & \cellcolor{good!18}82.8 \\
INT4, $dP{=}8$ & \cellcolor{cost!14}$5.28$ & \cellcolor{good!46}89.9 & \cellcolor{good!15}75.8 & \cellcolor{good!22}91.7 & \cellcolor{good!27}85.8 \\
INT4, $dP{=}32$ & \cellcolor{cost!31}$7.64$ & \cellcolor{good!50}91.2 & \cellcolor{good!23}78.7 & \cellcolor{good!22}91.7 & \cellcolor{good!31}87.2 \\
\bottomrule
\end{tabular}
\caption{Results on post-training state quantization on Qwen3.5-27B. Every row rounds the state to nearest with a group-$32$ scale. The prompt is rounded once per $256$-token kernel chunk throughout, so $dP$ varies the generation alone. \textbf{resident} is the effective bits stored for state during inference: the payload plus the decode buffer, at $0.098$ bits per value per token of period. \textbf{avg} is the unweighted mean of GSM8K, IFEval and RULER-VT; MMLU-Pro is not in it. Widths $4$ and $8$ were measured on NHR G\"ottingen and the widths between them here, both at batch $32$, which is the batch every row of this table shares.}
\label{tab:opening:27b}
\end{table*}

%% file: tabs/tab_opening_35b.tex
% generated by tabs/gen_opening.py -- do not edit by hand
\begin{table*}[t]
\centering\small
\setlength{\tabcolsep}{6pt}
\begin{tabular}{@{}lr|rrr|r@{}}
\toprule
\textbf{state} & \textbf{resident} & \textbf{GSM8K} & \textbf{IFEval} & \textbf{RULER-$32$k} & \textbf{avg} \\
\midrule
full precision & \cellcolor{cost!72}$32.00$ & \cellcolor{good!50}92.8 & \cellcolor{good!50}83.7 & \cellcolor{good!50}100.0 & \cellcolor{good!50}92.2 \\
INT8, $dP{=}1$ & \cellcolor{cost!39}$8.60$ & \cellcolor{good!41}90.5 & \cellcolor{good!9}69.3 & \cellcolor{good!5}83.0 & \cellcolor{good!15}80.9 \\
INT8, $dP{=}8$ & \cellcolor{cost!44}$9.28$ & \cellcolor{good!41}90.4 & \cellcolor{good!15}72.3 & \cellcolor{good!6}83.7 & \cellcolor{good!18}82.1 \\
INT8, $dP{=}32$ & \cellcolor{cost!61}$11.64$ & \cellcolor{good!41}90.4 & \cellcolor{good!14}72.1 & \cellcolor{good!6}84.0 & \cellcolor{good!18}82.2 \\
\addlinespace[2pt]
INT7, $dP{=}1$ & \cellcolor{cost!31}$7.60$ & \cellcolor{good!41}90.5 & \cellcolor{good!10}69.7 & \cellcolor{good!5}83.0 & \cellcolor{good!15}81.1 \\
INT7, $dP{=}8$ & \cellcolor{cost!36}$8.28$ & \cellcolor{good!39}89.8 & \cellcolor{good!12}71.2 & \cellcolor{good!5}83.0 & \cellcolor{good!16}81.3 \\
INT7, $dP{=}32$ & \cellcolor{cost!54}$10.64$ & \cellcolor{good!40}90.1 & \cellcolor{good!12}70.8 & \cellcolor{good!4}81.7 & \cellcolor{good!15}80.8 \\
\addlinespace[2pt]
INT6, $dP{=}1$ & \cellcolor{cost!24}$6.60$ & \cellcolor{good!35}88.9 & \cellcolor{good!4}65.6 & \cellcolor{good!5}83.0 & \cellcolor{good!11}79.2 \\
INT6, $dP{=}8$ & \cellcolor{cost!29}$7.28$ & \cellcolor{good!41}90.4 & \cellcolor{good!13}71.5 & \cellcolor{good!6}84.0 & \cellcolor{good!18}82.0 \\
INT6, $dP{=}32$ & \cellcolor{cost!46}$9.64$ & \cellcolor{good!41}90.4 & \cellcolor{good!12}70.8 & \cellcolor{good!6}83.3 & \cellcolor{good!16}81.5 \\
\addlinespace[2pt]
INT5, $dP{=}1$ & \cellcolor{cost!16}$5.60$ & \cellcolor{good!34}88.4 & \cellcolor{good!1}61.6 & \cellcolor{good!4}82.3 & \cellcolor{good!8}77.4 \\
INT5, $dP{=}8$ & \cellcolor{cost!21}$6.28$ & \cellcolor{good!39}89.8 & \cellcolor{good!11}70.4 & \cellcolor{good!4}82.3 & \cellcolor{good!15}80.9 \\
INT5, $dP{=}32$ & \cellcolor{cost!39}$8.64$ & \cellcolor{good!40}90.1 & \cellcolor{good!15}72.6 & \cellcolor{good!6}83.7 & \cellcolor{good!18}82.1 \\
\addlinespace[2pt]
INT4, $dP{=}1$ & \cellcolor{cost!9}$4.60$ & \cellcolor{good!12}80.0 & \cellcolor{good!0}54.7 & \cellcolor{good!0}77.3 & \cellcolor{good!1}70.7 \\
INT4, $dP{=}8$ & \cellcolor{cost!14}$5.28$ & \cellcolor{good!37}89.3 & \cellcolor{good!4}66.2 & \cellcolor{good!0}76.3 & \cellcolor{good!8}77.3 \\
INT4, $dP{=}32$ & \cellcolor{cost!31}$7.64$ & \cellcolor{good!43}90.9 & \cellcolor{good!8}68.8 & \cellcolor{good!0}75.7 & \cellcolor{good!10}78.4 \\
\bottomrule
\end{tabular}
\caption{Results on post-training state quantization on Qwen3.6-35B-A3B. Every row rounds the state to nearest with a group-$32$ scale. The prompt is rounded once per $256$-token kernel chunk throughout, so $dP$ varies the generation alone. \textbf{resident} is the effective bits stored for state during inference: the payload plus the decode buffer, at $0.098$ bits per value per token of period. \textbf{avg} is the unweighted mean of GSM8K, IFEval and RULER-VT; MMLU-Pro is not in it. Widths $4$ and $8$ were measured on NHR G\"ottingen and the widths between them here, both at batch $32$, which is the batch every row of this table shares.}
\label{tab:opening:35b}
\end{table*}

%% file: tabs/tab_permute.tex
% generated by tabs/gen_permute.py -- do not edit by hand
\begin{table*}[h]
\centering\small
\setlength{\tabcolsep}{3.5pt}
\begin{tabular}{@{}lrrrr@{\quad}rrrr@{}}
\toprule
& \multicolumn{4}{c}{\textbf{value/key max/RMS, $g{=}128$}} & \multicolumn{4}{c}{\textbf{outlier mass on columns}} \\
\cmidrule(lr){2-5}\cmidrule(lr){6-9}
\textbf{condition} & \textbf{4B} & \textbf{9B} & \textbf{27B} & \textbf{35B-A3B} & \textbf{4B} & \textbf{9B} & \textbf{27B} & \textbf{35B-A3B} \\
\midrule
original & $1.588$ & $1.555$ & $1.601$ & $1.871$ & $82.6$ & $78.6$ & $83.9$ & $87.8$ \\
$u_t$ channels permuted per token & $0.776$ & $0.760$ & $0.808$ & $0.963$ & $35.5$ & $34.5$ & $37.1$ & $45.4$ \\
$k$ entries permuted per token & $2.265$ & $2.220$ & $2.261$ & $2.580$ & $93.1$ & $92.3$ & $93.9$ & $92.7$ \\
\bottomrule
\end{tabular}
\caption{Permuting the channels of the delta value at every token gives key-axis groups the larger max/RMS; permuting the key entries widens the difference. Left: ratio of the value-axis to the key-axis max/RMS at $g{=}128$. Right: share of large entries that fall in the heaviest $5\%$ of columns, in percent.}
\label{tab:permute}
\end{table*}

%% file: tabs/tab_keybuf.tex
% generated by tabs/gen_keybuf.py -- do not edit by hand
\begin{table}[h]
\centering\scriptsize
\setlength{\tabcolsep}{2.5pt}
\begin{tabular}{@{}lrrrrrrrr@{}}
\toprule
& & \multicolumn{4}{c}{\textbf{IFEval}} & \multicolumn{3}{c}{\textbf{RULER-$32$k}} \\
\cmidrule(lr){3-6}\cmidrule(lr){7-9}
$k$/$v$ & \textbf{bits} & 4B & 9B & 27B & 35B & 9B & 27B & 35B \\
\midrule
full precision & $32.00$ & 83.2 & 84.5 & 86.9 & 83.4 & 100.0 & 100.0 & 100.0 \\
\midrule
\multicolumn{9}{@{}l}{\emph{key axis}} \\
\quad int$8$ / int$8$ & $6.44$ & 81.5 & 78.2 & 79.9 & 72.8 & 98.3 & 95.0 & 84.3 \\
\quad int$4$ / bf$16$ & $6.82$ & 78.2 & 82.4 & 86.5 & 80.4 & 100.0 & 100.0 & 100.0 \\
\quad bf$16$ / int$4$ & $6.44$ & 13.9 & 22.7 & 37.0 & 12.2 & 1.7 & 34.3 & 0.0 \\
\midrule
\multicolumn{9}{@{}l}{\emph{value axis}} \\
\quad int$8$ / int$8$ & $6.44$ & 80.4 & 76.7 & 78.7 & 71.5 & 97.0 & 94.0 & 82.7 \\
\quad int$4$ / bf$16$ & $6.82$ & 77.1 & 81.1 & 84.8 & 80.8 & 100.0 & 96.3 & 99.7 \\
\quad bf$16$ / int$4$ & $6.44$ & 14.0 & 22.7 & 37.5 & 11.3 & 4.3 & 32.3 & 0.0 \\
\bottomrule
\end{tabular}
\caption{Rounding the buffered values to four bits reduces accuracy more than rounding the buffered keys. PACE-4/8 state quantizer at $dP{=}8$; $k$/$v$ gives the widths of the buffered keys and values; bits are resident bits per value.}
\label{tab:keybuf}
\end{table}